%% file: main.tex
\documentclass[letterpaper]{article} 
\usepackage{aaai25}  
\usepackage{times}  
\usepackage{helvet}  
\usepackage{courier}  
\usepackage[hyphens]{url}  
\usepackage{graphicx} 
\usepackage{natbib}  
\usepackage{caption} 
\usepackage{algorithm}
\usepackage{algorithmic}
\usepackage{xcolor}
\usepackage{newfloat}
\usepackage{listings}
\DeclareCaptionStyle{ruled}{labelfont=normalfont,labelsep=colon,strut=off} 
\floatstyle{ruled}
\newfloat{listing}{tb}{lst}{}
\floatname{listing}{Listing}
\usepackage{kotex}
\usepackage{comment}
\usepackage{amsmath}
\usepackage{amssymb}
\usepackage{booktabs}
\usepackage{multirow}
\usepackage{makecell}

\newcommand{\tabref}[1]{Table~\ref{#1}}
\newcommand{\equref}[1]{Equation~\ref{#1}}
\newcommand{\figref}[1]{Figure~\ref{#1}}

\title{Intrinsic Image Decomposition for Robust Self-supervised Monocular Depth Estimation on Reflective Surfaces}
\author {
    Wonhyeok Choi\equalcontrib,
    Kyumin Hwang\equalcontrib,
    Minwoo Choi,
    Kiljoon Han,\\
    Wonjoon Choi,
    Mingyu Shin,
    Sunghoon Im\thanks{S. Im is the corresponding author.}
}
\affiliations {
    Daegu Gyeongbuk Institute of Science and Technology, South Korea\\
    \{smu06117, kyumin, subminu, kiljoon\_h, wjchoi, alsrb4446, sunghoonim\}@dgist.ac.kr
}

\begin{document}

\maketitle

\begin{abstract}
Self-supervised monocular depth estimation (SSMDE) has gained attention in the field of deep learning as it estimates depth without requiring ground truth depth maps. This approach typically uses a photometric consistency loss between a synthesized image, generated from the estimated depth, and the original image, thereby reducing the need for extensive dataset acquisition. However, the conventional photometric consistency loss relies on the Lambertian assumption, which often leads to significant errors when dealing with reflective surfaces that deviate from this model. To address this limitation, we propose a novel framework that incorporates intrinsic image decomposition into SSMDE. Our method synergistically trains for both monocular depth estimation and intrinsic image decomposition. The accurate depth estimation facilitates multi-image consistency for intrinsic image decomposition by aligning different view coordinate systems, while the decomposition process identifies reflective areas and excludes corrupted gradients from the depth training process. Furthermore, our framework introduces a pseudo-depth generation and knowledge distillation technique to further enhance the performance of the student model across both reflective and non-reflective surfaces. Comprehensive evaluations on multiple datasets show that our approach significantly outperforms existing SSMDE baselines in depth prediction, especially on reflective surfaces.

\end{abstract}

%

\input{1_introduction}

\input{2_related_work}

\input{3_method}

\input{4_experiments}

\section{Conclusion}

This paper presents the first end-to-end approach that leverages intrinsic image decomposition to address the challenges posed by reflective surfaces, which are known to compromise depth accuracy in SSMDE.
We introduce a new method for localizing the pixel-level reflective regions without the need for additional labeled annotation by using estimated intrinsic components.
Building on this, our reflection-aware SSMDE method leverages the localized information to significantly improve depth accuracy.
It prevents erroneous depth learning caused by these regions violating the Lambertian assumption.
Experimental results demonstrate that our method significantly outperforms conventional self-supervised approaches across the various indoor datasets consisting of numerous reflective surfaces.
Moreover, the proposed method achieves competitive performance compared to the method leveraging 3D knowledge distillation, which requires excessive training costs.

\section{Acknowledgements}

This work was supported by Korea Research Institute for defense Technology planning and advancement through Defense Innovation Vanguard Enterprise Project, funded by Defense Acquisition Program Administration (R230206) and the National Research Foundation of Korea (NRF) grant funded by the Korea government (MSIT) (No. RS-2023-00210908).

\bibliography{aaai25}

\clearpage
\input{5_appendix}

\end{document}

%% file: 1_introduction.tex
\section{Introduction}
Self-supervised monocular depth estimation (SSMDE) is a method to estimate depth using only sequential images during training, without the need for ground truth depth maps.
This method has significantly expanded its applicability in commercial datasets, eliminating the need for expensive data acquisition methods like LiDAR. 
The key to the success of SSMDE lies in its ability to synthesize one image into another using estimated depth from sequential images and enforcing photometric consistency constraints.
Over the years, SSMDE has consistently improved and is increasingly being applied in industries such as autonomous driving, robotics, and mixed reality.

\begin{figure}[t]
\centering
\includegraphics[width=0.9\columnwidth]{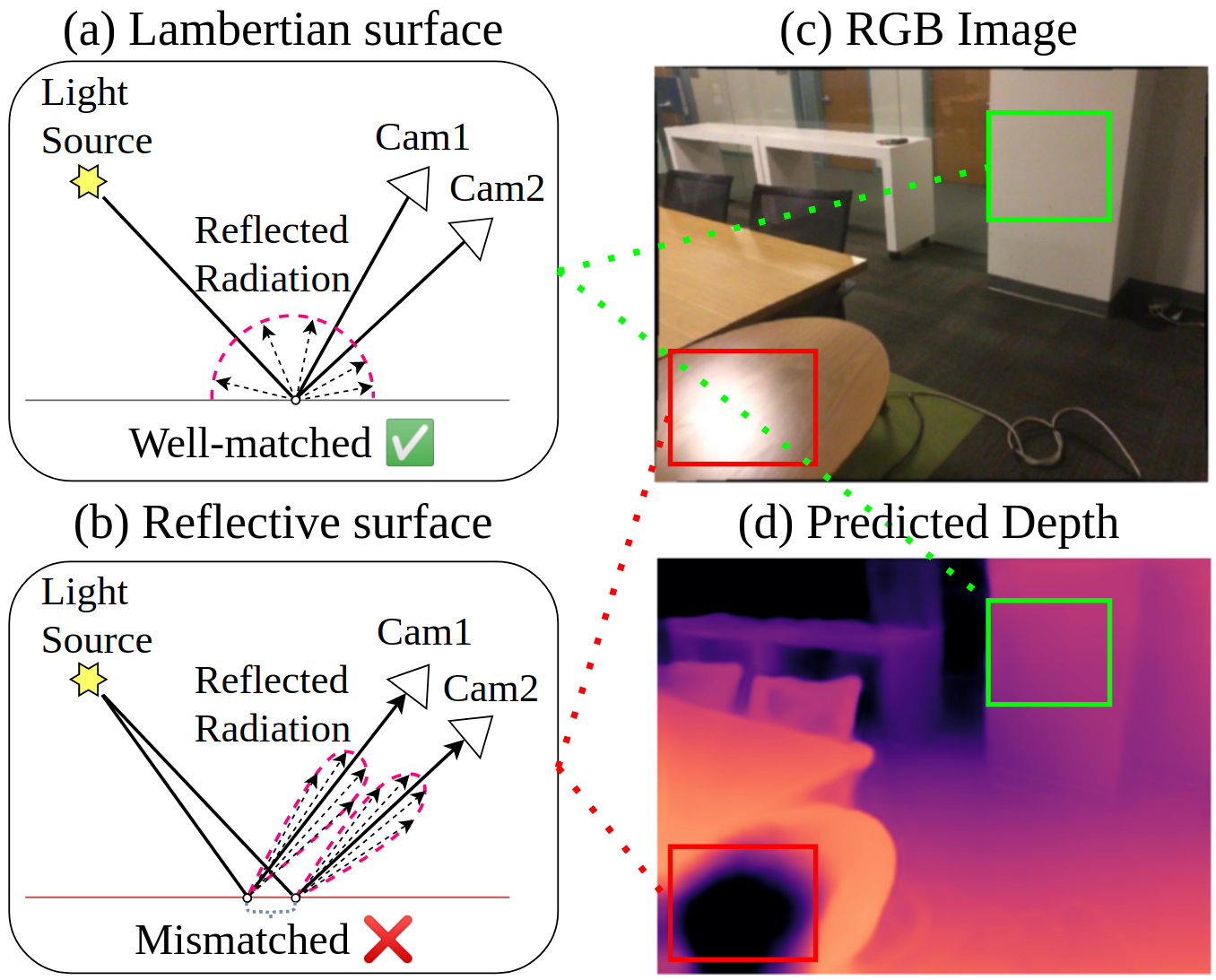} 
\caption{Illustrations of violations of the photometric constancy principle on non-Lambertian surfaces.}
\label{fig:non_lambertian}
\vspace{-5pt}
\end{figure}

Recently, the focus of SSMDE research has shifted from merely enhancing depth estimation performance to investigating intrinsic network issues, particularly the generalization of trained models to changes in domain~\citep{saunders2023self,gasperini2023robust,bae2023study}. 
Some approaches have highlighted a critical issue: SSMDE often relies on the Lambertian assumption as a learning strategy, which does not hold true for non-Lambertian surfaces like mirrors and reflective materials commonly found in indoor scenes.
These surfaces violate the photometric constancy principle, a cornerstone assumption of SSMDE, significantly degrading depth quality as shown in \figref{fig:non_lambertian}.
To address these issues, recent methods have been developed to localize the regions that violate these assumptions using additional segmentation masks~\citep{costanzino2023learning}, based on the uncertainty of multiple networks~\citep{shi20233d}, and to selectively refine the depth in these areas using techniques such as inpainting and 3D mesh rendering.
However, the computational intensity of 3D mesh rendering and ensemble uncertainty, along with the requirement for additional annotations, diminish the advantages of self-supervised approaches.

To address these challenges, we propose a novel end-to-end, plug-and-play method that leverages intrinsic image decomposition, specifically utilizing the intrinsic residual model~\citep{tominaga1994dichromatic, maxwell2008bi}. 
This method decomposes an image into diffuse and residual components, the latter of which often varies with color or luminosity relative to the camera viewpoint—common complication in training SSMDEs.
Our method specifically addresses these variations by localizing per-pixel reflective regions.
It compares the photometric error between the diffuse image, where the residual component has been removed, and the original RGB image. 
Reflective regions are identified through this comparison and then excluded from the depth learning process.
By systematically excluding these problematic reflective regions, our method significantly improves depth accuracy.



To demonstrate the effectiveness of the proposed method, we conduct evaluations using three representative SSMDE baselines~\citep{godard2019digging, lyu2021hr, zhao2022monovit} across three datasets~\citep{dai2017scannet, shotton2013scene, ramirez2023booster}. These datasets include various realistic indoor scenes comprising numerous non-reflective and reflective surfaces.
The results demonstrate significant improvements in depth prediction accuracy, particularly on reflective and non-Lambertian surfaces.
Our contributions are threefold as follows:
\begin{itemize}
\item To the best of our knowledge, we are the first to introduce a novel SSMDE that employs the self-supervised intrinsic image decomposition for robust depth estimation. 
\item We present a method for identifying reflective regions using dynamic viewpoint monocular image sequences without requiring explicitly labeled data.
\item We propose a reflection-aware depth training process that enhances the accuracy of depth prediction in our SSMDE framework.

\end{itemize}

%% file: 2_related_work.tex
\section{Related Work}

\subsection{Self-supervised Monocular Depth Estimation}
Monodepth~\citep{godard2017unsupervised}, a pioneering works in self-supervised monocular depth estimation, minimizes the photometric error by synthesizing a left image from a right image using stereo geometry.
Following this work, interest in SSMDE has surged, leading to significant developments using CNN models~\citep{zhou2017unsupervised,godard2019digging,guizilini20203d,lyu2021hr,zhou2021r} and ViT models~\citep{zhao2022monovit, bae2023deep, wang2024sqldepth,zhang2023lite}.

In addition to structural advancements, substantial efforts have been made to address inherent challenges in SSMDE, such as scale-ambiguity~\citep{guizilini20203d,guizilini2022full}, out-of-distribution generalization~\citep{saunders2023self, gasperini2023robust,bae2023study}. 
Recent works~\citep{costanzino2023learning,shi20233d} have tackled the degradation of depth accuracy in non-Lambertian regions, a common issue in models trained under photometric consistency constraints.
These works leverage additional resources such as ensemble uncertainty~\citep{shi20233d} or segmentation mask annotations~\citep{costanzino2023learning} to identify the non-Lambertian regions.
They refine the depth estimates of these regions using techniques like inpainting-based color augmentation or 3D mesh rendering.

\subsection{Intrinsic Image Decomposition}
Intrinsic image decomposition is a deeply researched field with a history spanning nearly five decades.
Early approaches~\citep{land1971lightness, shen2008intrinsic, zhao2012closed, rother2011recovering, shen2011intrinsic, garces2012intrinsic} utilize prior knowledge, such as the assumption that large image gradients correspond to changes in albedo, to perform this task through non-learning-based method optimization.
With the advent of labeled datasets, the focus has shifted towards learning-based methods~\citep{barron2014shape, narihira2015direct, shi2017learning, kim2016unified}, which primarily rely on ground truth intrinsic components.
Some recent state-of-the-art learning-based intrinsic decomposition methods~\citep{careaga2023intrinsic, luo2024intrinsicdiffusion, zeng2024rgb} have demonstrated impressive performance, enabling natural and high-quality image rendering. 

On the other hand, some studies~\citep{weiss2001deriving, sunkavalli2007factored, laffont2012coherent, laffont2015intrinsic, duchene2015multi, ma2018single, yu2019inverserendernet} have proposed self-supervised methods for training networks using only RGB image sequences without ground-truth information on intrinsic components.
These studies primarily utilize image sequences with a fixed viewpoint and a static scene, but with varying illumination, such as time-lapse videos or synthetic image sequences. 
They exploit the assumption that the albedo remains constant in such static scenes to conduct the learning process.
Notably, \citet{yu2019inverserendernet} is the first to use the multi-view image sequences but this method still relies on the depth annotations to obtain the pre-trained multi-view stereo network.
To the best of our knowledge, our method is the first fully self-supervised framework that does not require any explicitly labeled data while leveraging dynamic viewpoint monocular image sequences.


%% file: 3_method.tex
\section{Method}

\begin{figure}[t]
\centering
\includegraphics[width=0.95\columnwidth]{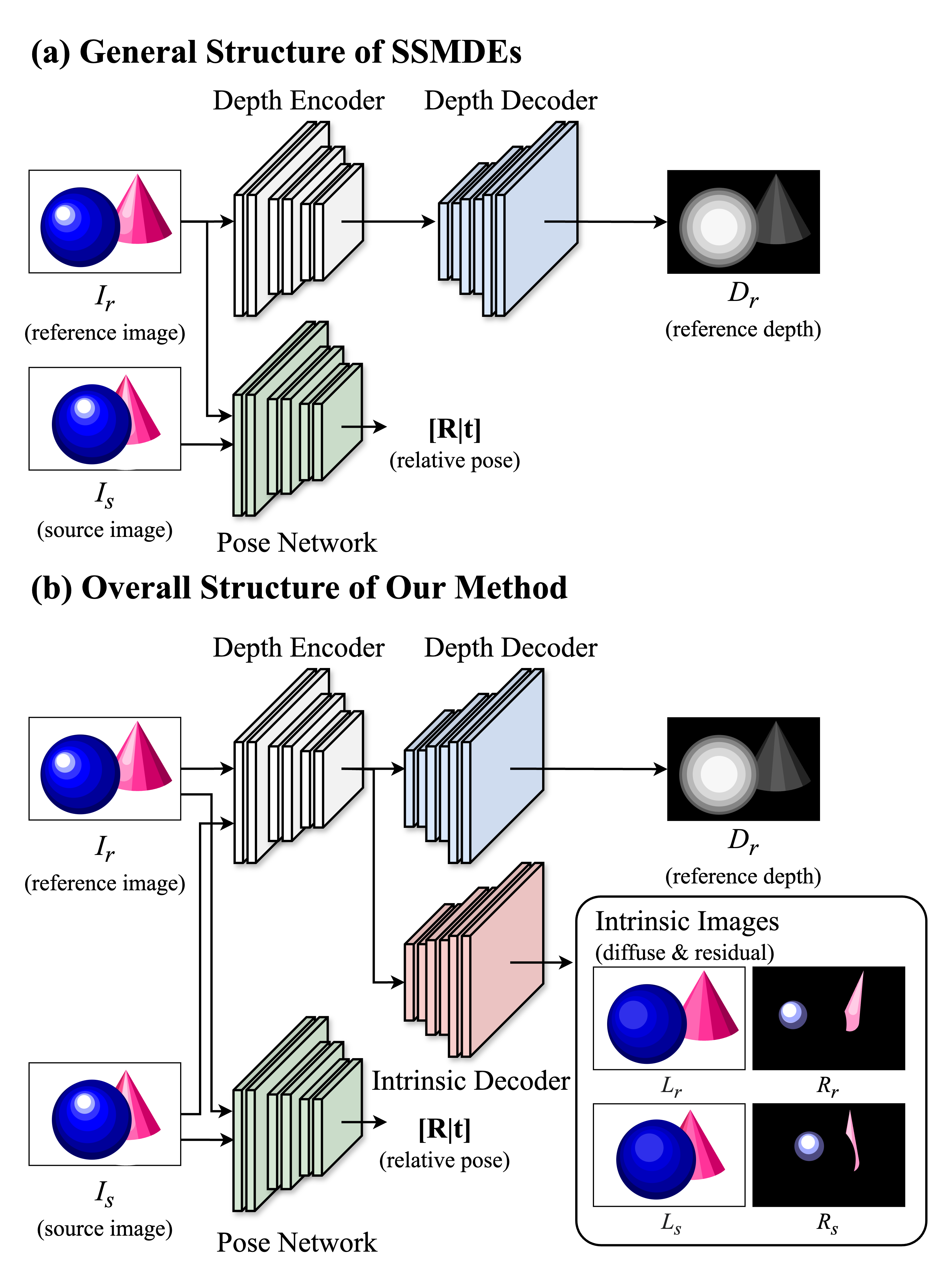} 
\caption{Overall structure of our method.}
\label{fig:overall_pipeline}
\vspace{-5pt}
\end{figure}



Our framework consists of two primary branches: the intrinsic image decomposition branch and the depth estimation branch as shown in \figref{fig:overall_pipeline}.
These branches collaboratively estimate intrinsic components and depth from RGB images, respectively. 
We replicate the architecture of the depth decoder in the intrinsic image decoder to additionally predict the intrinsic images.
Our method specifically aims to enhance the depth prediction accuracy in the reflective regions, which are problematic for depth training, by leveraging the intrinsic components.

\subsection{Preliminary}
\subsubsection{Basic assumption of general SSMDEs}
To train the monocular depth estimation network in a self-supervised manner using photometric loss, conventional works typically assume that the following three conditions must be satisfied:
\begin{itemize}
    \item[\textit{\textbf{a1}}] The scene being modeled is static while camera is in motion~\citep{godard2019digging, zhou2017unsupervised}.
    \item[\textit{\textbf{a2}}] Objects are perceived to maintain a consistent level of brightness across different views~\citep{godard2019digging}.
    \item[\textit{\textbf{a3}}] All surfaces in the scene are Lambertian~\citep{godard2019digging, zhou2017unsupervised}.
\end{itemize}
Nevertheless, outdoor scene datasets contain elements that violate these assumptions, such as moving vehicles or pedestrians, rapidly changing weather conditions.
Some SSMDE methods use an auto-mask~\citep{godard2019digging} to exclude pixel-level regions that violate the static scene assumption (\textit{\textbf{a1}}).
While the Lambertian surface assumption (\textit{\textbf{a3}}) is not universally valid, it still provides a reasonable approximation for most outdoor scenes.
However, this assumption (\textit{\textbf{a3}}) fails notably in indoor environments where reflective surfaces, mirrors, and transparent objects can severely harm the depth accuracy due to the diverse object types and lighting conditions.

\subsubsection{General self-supervised intrinsic image decomposition}
Self-supervised intrinsic image decomposition methods typically use time-lapse videos or synthetic datasets comprising image sequences under varying lighting conditions. These works aim to decompose an RGB image $I$ into its albedo $A$ and shading $S$ components as follows:
\begin{align}
\begin{split}
\label{equ:intrinsic_diffuse}
I = A \odot S,
\end{split}
\end{align}
where $\odot$ represents the element-wise multiplication operator.
Given that albedo is inherently invariant to illumination changes, it remains consistent across all images within the same sequence.
This consistency in albedo allows these methods to effectively decompose the intrinsic components of the images.

\subsection{Training of Intrinsic Image Decomposition Branch}

\subsubsection{Our formulation of intrinsic decomposition}
The intrinsic diffuse model in \equref{equ:intrinsic_diffuse} cannot be directly applied to a scenario of SSMDE due to these problems:
\begin{itemize}
    \item[\textit{\textbf{p1}}] The lighting conditions in the image sequence are static ($\because$ \textit{\textbf{a2}}).
    \item[\textit{\textbf{p2}}] The camera coordinates within the image sequence are unaligned ($\because$ \textit{\textbf{a1}}).
    \item[\textit{\textbf{p3}}] The intrinsic diffuse model assumes all surfaces in the scene are Lambertian.
\end{itemize}
To address these challenges, we newly design the training method, facilitating the learning of intrinsic image decomposition in SSMDE scenarios, where only unaligned RGB image sequences are available.

To align the camera coordinate of multiple images in a sequence, we use the view-synthesized method~\cite{godard2019digging}, which is generally adopted in SSMDEs.
We also reformulate the intrinsic image decomposition to decompose the diffuse and residual, which are the view-independent and view-dependent components, respectively.
Instead of using the intrinsic diffuse model, we adopt the intrinsic residual model~\citep{tominaga1994dichromatic, maxwell2008bi}, which accommodates the specular reflections that vary with the camera view direction. 
We formulate the residual model in logarithmic space, where the RGB image exhibits shadow and illumination-invariant chromacity~\citep{maxwell2008bi}. This allows for the decomposition of the diffuse image and residual image from the original image as follows:
\begin{align}
\begin{split}
\label{equ:our_formulation}
\log I = \log L + \log R,
~\text{where}~L = A \odot S,
\end{split}
\end{align}
where $L$ is the diffuse image, defined as the element-wise multiplication of albedo and shading images. The diffuse component remains constant irrespective of the camera view direction. $R$ is the residual component varying depending on the camera view direction.

\subsubsection{Loss term of intrinsic network}
The objective of the intrinsic image decomposition network is to appropriately decompose the diffuse images and residual images from a given reference image $I_r$, a source image $I_s$ (or source images), relative camera pose between reference view and source view $[\mathbf{R|t}]$, and their camera intrinsic $\mathbf{K}$.
The network is designed to predict the diffuse component $L_r$ and the residual component $R_r$ from the reference image $I_r$. Similarly, it predicts the source diffuse $L_s$ and residual $R_s$ from the source image $I_s$.
The predicted intrinsic components are required to flawlessly reconstruct the original RGB image by \equref{equ:our_formulation}.
Accordingly, we design the reconstruction loss $\mathcal{L}_{\text{recon}}$ based on this formulation as follows:
\begin{align}
\begin{split}
\label{equ:recon_loss}\
\mathcal{L}_{\text{recon}} = |\log(I_r) - \log(L_r) - \log(R_r)|_1.
\end{split}
\end{align}

Corresponding surface points in diffuse images must have the same color and luminance because diffuse components do not vary with the camera view direction.
To enable the model to leverage the diffuse consistency with static camera coordinates, the model aligns the camera coordinate of the source RGB image and diffuse image $\Gamma \in \{I, L\}$ to the reference camera coordinate by using view synthesis~\citep{godard2019digging} as follows:
\begin{equation}
\begin{gathered}
    \label{equ:warping}
    \Gamma_{s2r}[u,v] = \Gamma_s[\hat{u},\hat{v}],~\Gamma \in \{I, L\},\\
    (\hat{u},\hat{v}) = \text{proj}\bigg({\mathbf{K}}{[\mathbf{R|t}]}_{r2s} D_{r}[u,v] \mathbf{K}^{-1}
    \begin{pmatrix}
    u \\ v \\ 1
    \end{pmatrix}
    \bigg),
\end{gathered}
\end{equation}
where $(u, v)$ and $(\hat{u}, \hat{v})$ represent the corresponding indices of image coordinates in the warping process. The function proj$(\cdot)$ serves as the projection function that maps homogeneous coordinates to image coordinates. Additionally, $D_{r}$ is the reference depth prediction from the depth branch.
If the depth prediction $D_r$ is sufficiently accurate, the aligned diffuse image $L_r$ and $L_{s2r}$ should be identical, reflecting diffuse consistency. To leverage this property, we design the cross-reconstruction loss $\mathcal{L}_{\text{cross}}$ as follows:
\begin{align}
\begin{split}
\label{equ:cross_loss}
    \mathcal{L}_{\text{cross}} = |\log(I_{r}) - \log(L_{s2r}) - \log(R_{r})|_1.
\end{split}
\end{align}
The proposed loss terms $\mathcal{L}_{\text{recon}}$ and $\mathcal{L}_{\text{cross}}$ are designed to ensure that the outputs of intrinsic model adhere to the diffuse-residual model formulation and diffuse consistency.

However, these constraints allow for numerous optimal solutions, including some that are unsuitable.
For example, an outcome where the diffuse image is a plain white image and the residual image matches the original RGB image would reduce these loss terms to zero, yielding nonsensical intrinsic components.
To ensure that the model does not provide such abnormal intrinsic images, we design additional contrastive loss term $\mathcal{L}_{\text{cts}}$ that encourages the model to decompose a different diffuse image for each image sequence as follows:
\begin{align}
\begin{split}
\mathcal{L}_{\text{cts}} = \sum_{i=0}^{b}\sum_{\substack{j=0 \\ i \neq j}}^{b} \max(\delta - \|L_{s2r}^{i} - L_{r}^{j}\|_2,~0),
\end{split}
\end{align}
where $b$ is the batch size and $\delta$ is the proper margin to regulate the distance between diffuse images from different images.
The total loss for intrinsic image decomposition $\mathcal{L}_{\text{itr}}$ is formed as follows:
\begin{align}
\begin{split}
    \mathcal{L}_{\text{itr}} = \lambda_{\text{recon}} \mathcal{L}_{\text{recon}} + \lambda_{\text{cross}} \mathcal{L}_{\text{cross}} + \lambda_{\text{cts}} \mathcal{L}_{\text{cts}}.
\end{split}
\end{align}

\subsection{Training of Depth Estimation Branch}
Most SSMDEs use a combination of an L1 and single scale SSIM~\citep{wang2004image} as the photometric error, which is proposed by~\citet{godard2017unsupervised}:
\begin{align}
\begin{split}
    \mathcal{P}(\Gamma_{r}, \Gamma_{s2r}) = M \odot \Bigl(\alpha \bigl(1-\text{SSIM}(\Gamma_{r}, \Gamma_{s2r})\bigl)/2 \\ 
    + (1-\alpha)\|\Gamma_{r} - \Gamma_{s2r}\|_1 \Bigl),~\Gamma \in \{I, L\},
\end{split}
\end{align}
where $\Gamma_{s2r}$ is a set that includes warped image or diffuse component $\{I_{s2r}, L_{s2r}\}$ according to \equref{equ:warping}. Here, $M$ is the auto-mask~\citep{godard2019digging}, and $\alpha$ is a parameter within $[0, 1]$ balancing the contribution of the two loss terms.

While general SSMDEs minimize the photometric error $\mathcal{P}(I_r, I_{s2r})$, the mask $M$ is employed to selectively filter out the gradients from regions affected by moving objects or static camera movements, addressing the assumption \textit{\textbf{a1}}. However, this method does not account for the Lambertian assumption \textit{\textbf{a3}}.
To identify for regions that substantially violate the assumption \textit{\textbf{a3}}, we compute the photometric error for both aligned RGB pairs and diffuse pairs as follows:
\begin{equation}
\begin{gathered}
    E_{I} = \mathcal{P}(I_r, I_{s2r}),~
    E_{L} = \mathcal{P}(L'_{r}, L'_{s2r}),\\
    L' = \exp(\log(I) - \log(R))~~\text{($\because$ \equref{equ:our_formulation})},
\end{gathered}
\end{equation}
where $E_{I}$ and $E_{L}$ are the photometric errors between RGB pairs and diffuse pairs, respectively.
Note that we use the pseudo-diffuse $L'$ computed from the RGB image and the predicted residual image rather than the predicted diffuse image $L$ due to the leak of high-frequency details during image-to-image translation.

Since the residual image varies with color or luminosity relative to the camera viewpoint, photometric errors can arise from the residual component.
As a result, the photometric error between a pair of diffuse images, where the residual components have been removed from the color images, is lower than the photometric error between the original color images.
We use the Mahalanobis distance~\citep{mahalanobis1936mahalanobis} to pinpoint and exclude problematic pixel-level regions from the training where the difference between the image error and diffuse error in the photometric error space is significant as follows:
\input{tables/table1}
\begin{align}
\begin{split}
\label{equ:reflection_mask}
    &M_{R}[u, v] = 
    \begin{cases}
    0, \text{ if }z_{L}[u,v] < z_{I}[u,v]\\
    1, \text{ else }
    \end{cases},~\text{where}\\
    z_{I} &= \sum_{u,v}\sqrt{(E_{I}[u,v]-\mu_{I})\mathbf{\sigma}_{I}^{-1}(E_{I}[u,v]-\mu_{I})^T},\\
    z_{L} &= \sum_{u,v}\sqrt{(E_{L}[u,v]-\mu_{L})\sigma_{L}^{-1}(E_{L}[u,v]-\mu_{L})^T},
\end{split}
\end{align}
where $\{\mu_{I}, \mu_{L}\}$ represent the mean vectors and $\{\sigma_{I}, \sigma_{L}\}$ denote the covariance vector of per-pixel vectors from the photometric error of RGB and diffuse images, respectively. $M_{R}$ is a per-pixel binary mask, similar in function to the auto-mask $M$, designed to exclude non-diffuse regions. 
Importantly, the Mahalanobis distance is employed as a metric to assess the divergence between a sample point and a distribution, allowing the network to selectively identify and filter out specular regions.
This is achieved by recognizing areas where the photometric error significantly decreases following the removal of the residual component in specific pixels.

Similar to the auto-masking scheme, our method mitigates the influence of the attenuated gradients from reflective regions during training.
This is achieved by performing element-wise multiplication of the photometric error with the obtained mask $M_R$ as follows:
\begin{align}
\begin{split}
    \mathcal{L}_{\text{depth}} = M_{R} \odot \mathcal{P}(I_r, I_{s2r}).
\end{split}
\end{align}

Lastly, our framework is designed to simultaneously learn intrinsic image decomposition and depth estimation by summing the losses $\mathcal{L}_{\text{itr}}$ and $\mathcal{L}_{\text{depth}}$ as follows:
\begin{align}
\begin{split}
    \mathcal{L}_{\text{total}} = \mathcal{L}_{\text{itr}} + \mathcal{L}_{\text{depth}}.
\end{split}
\end{align}

%% file: tables/table1.tex
\begin{table*}[t!]

  \begin{center}
  \resizebox{0.8\textwidth}{!}{
  \begin{tabular}{@{}c|lccccccc@{}}
  \toprule
    Dataset & Method & Abs Rel $\downarrow$ & Sq Rel $\downarrow$ & RMSE $\downarrow$ & RMSE log $\downarrow$ & $\delta < 1.25 \uparrow$ & $\delta < 1.25^2 \uparrow$ & $\delta < 1.25^3 \uparrow$ \\
    \midrule
    \multirow{6}{*}{\rotatebox[origin=c]{90}{\makecell{ScanNet-Refl.\\Test set}}} & Monodepth2 & 0.181 & 0.160 & 0.521 & 0.221 & \textbf{0.758} & 0.932 & 0.976 \\
    & Monodepth2 + \textit{Ours} & \textbf{0.163} & \textbf{0.103} & \textbf{0.472} & \textbf{0.205} & 0.757 & \textbf{0.940} & \textbf{0.986} \\
    \cmidrule{2-9}
    & HRDepth & 0.182 & 0.168 & 0.530 & 0.225 & 0.749 & \textbf{0.937} & 0.979 \\
    & HRDepth + \textit{Ours} & \textbf{0.165} & \textbf{0.104} & \textbf{0.471} & \textbf{0.206} & \textbf{0.752} & 0.934 & \textbf{0.984} \\
    \cmidrule{2-9}
    & MonoViT & 0.154 & 0.129 & 0.458 & 0.197 & \textbf{0.822} & 0.955 & 0.979 \\
    & MonoViT + \textit{Ours} & \textbf{0.142} & \textbf{0.093} & \textbf{0.421} & \textbf{0.180} & 0.815 & \textbf{0.957} & \textbf{0.985} \\
  \bottomrule
  \toprule
    \multirow{6}{*}{\rotatebox[origin=c]{90}{\makecell{ScanNet-Refl.\\Val. set}}} & Monodepth2 & 0.206 & 0.227 & 0.584 & 0.246 & 0.750 & 0.912 & 0.961 \\
    & Monodepth2 + \textit{Ours} & \textbf{0.158} & \textbf{0.100} & \textbf{0.462} & \textbf{0.200} & \textbf{0.769} & \textbf{0.939} & \textbf{0.986} \\
    \cmidrule{2-9}
    & HRDepth & 0.213 & 0.244 & 0.605 & 0.255 & 0.741 & 0.906 & 0.961 \\
    & HRDepth + \textit{Ours} & \textbf{0.160} & \textbf{0.102} & \textbf{0.463} & \textbf{0.201} & \textbf{0.773} & \textbf{0.939} & \textbf{0.986} \\
    \cmidrule{2-9}
    & MonoViT & 0.179 & 0.206 & 0.557 & 0.227 & 0.819 & 0.930 & 0.963 \\
    & MonoViT + \textit{Ours} & \textbf{0.139} & \textbf{0.106} & \textbf{0.446} & \textbf{0.179} & \textbf{0.840} & \textbf{0.953} & \textbf{0.984} \\
  \bottomrule
  \end{tabular}}
  \end{center}
  \caption{Main results on the ScanNet-Reflection Test and Validation sets.}
  \label{tab:reflection}
\end{table*}

%% file: 4_experiments.tex
\section{Experiments}
\label{sec:experiments}

\subsection{Baselines and Training Setups}
We incorporate experiments with three leading architectures that have demonstrated outstanding performance in previous research: Monodepth2~\citep{godard2019digging}, HRDepth~\citep{lyu2021hr}, and MonoViT~\citep{zhao2022monovit}.
Each baseline was trained in a self-supervised manner using the photometric error minimization strategy initially proposed by Monodepth2, and they are referred to using the general label [Baseline].
To demonstrate the effectiveness of our method relative to these baselines, we adapt the decoder of each baseline architecture to our training protocol, incorporating intrinsic networks. This modified training approach is denoted as [Baseline + Ours].

\input{tables/table2}

\begin{figure*}[t!]
\centering
\vspace{-9pt}
\includegraphics[width=0.9\textwidth]{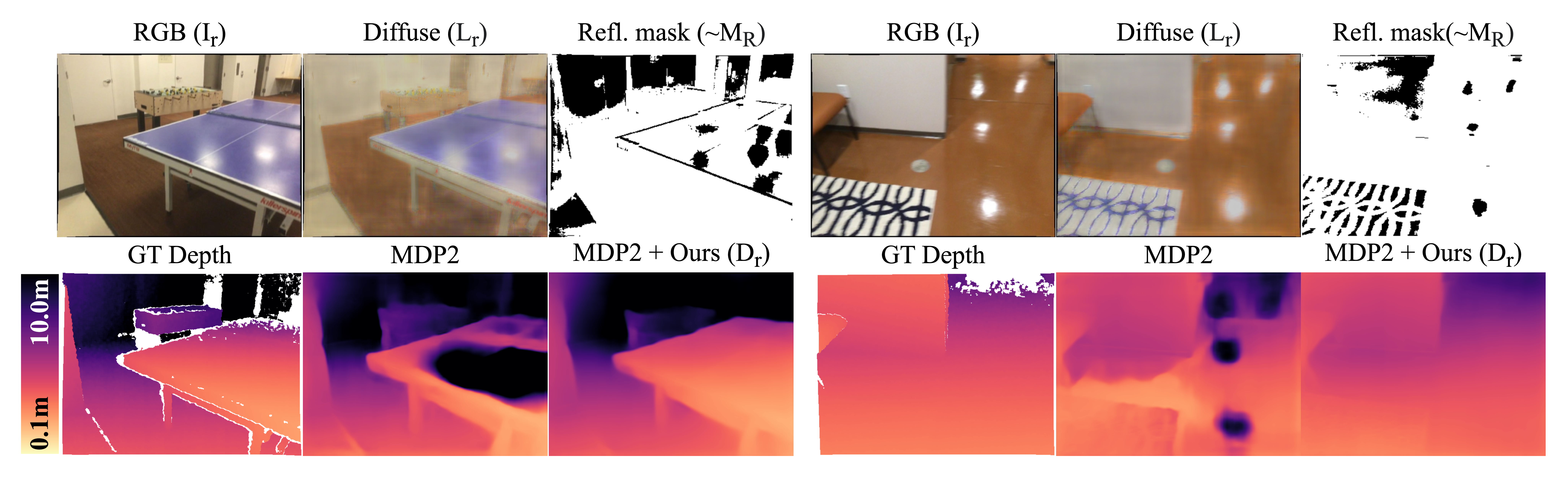} 
\caption{Qualitative results between monodepth2 (denoted as MDP2) and our method.}
\label{fig:qualitative}
\vspace{-1pt}
\end{figure*}

\subsection{Implementation Details}
The batch size for training is configured differently across the architecture: 12 for Monodepth2 and HRDepth, and 8 for MonoViT.
All networks start with a learning rate of 1e-4, which is decreased by a factor of 10 following the 26th and 36th epochs, within a total training duration of 41 epochs. 
The balancing weights for photometric error and smoothness loss follow the policy established by Monodepth2. 
For resolution, a training dimension of 384$\times$288 is utilized, and no post-processing techniques, such as median-scaling, are employed in any of the quantitative evaluations.
Additionally, similar to 3D Distillation~\citep{shi20233d}, camera poses are assumed to be known during training.
Depth boundaries were set between a minimum of 0.1m and a maximum of 10m.
Weight balancing parameters for intrinsic branch $\{\lambda_{\text{recon}}, \lambda_{\text{cross}}, \lambda_{\text{cts}}\}$ are set at $\{1.0, 1.0, 0.01\}$ respectively, with $\delta$ for $\lambda_{\text{cts}}$ fix at 5.0. For the quantitative evaluation of depth estimation, we utilized metrics that are widely recognized and standard within the field~\citep{eigen2014depth, geiger2012we}. 

\subsection{Datasets}
\begin{itemize}
    \item \textbf{\textit{ScanNetv2}}~\citep{dai2017scannet} consists of indoor scenes captured by an RGB-D scanner, including 1,201 training scenes, 312 validation scenes, and 100 test scenes.
    To simulate training scenarios dominated by non-Lambertian surfaces and to ensure fair comparison with 3D Distillation~\citep{shi20233d}, we utilize the training triplet splits provided by the authors, encompassing 45,539 training images, 439 validation images, and 121 test images.
    \item \textbf{\textit{7-Scenes}}~\citep{shotton2013scene} comprises seven indoor scenes and has been extensively utilized in 3D reconstruction and camera localization studies.
    To evaluate the cross-dataset generalization of the proposed method, we follow the evaluation splits specified by~\citet{long2021multi, bae2022multi}.
    \item \textbf{\textit{Booster}}~\citep{ramirez2023booster} comprises indoor scenes with a variety of non-Lambertian surfaces, including transparent, mirrored, and specular surfaces.
    In line with the work of~\citet{costanzino2023learning}, who evaluate the performance of monocular and stereo depth networks on non-Lambertian surfaces, we adopt their training split as our test split to evaluate our method.
\end{itemize}

\subsection{Performance Evaluation}
\subsubsection{ScanNet-Reflection dataset}
To demonstrate the effectiveness of the proposed method on reflective surfaces, we measure depth accuracy using the Scannet-Reflection test and validation sets, which specifically include scenes with reflective surfaces.
The results indicate that all baselines demonstrate consistent performance improvements across nearly all evaluated metrics.
Notably, the \textit{Abs rel} metric display significant gains, with an average improvement of 9.02\% on the test set and an impressive 23.51\% on the validation set, as shown in \tabref{tab:reflection}.
This indicates the proposed method encourages the network to accurately learn depth in reflective surfaces, as depicted in \figref{fig:qualitative}.

\subsubsection{ScanNet-NoReflection dataset}
Since the proposed method pinpoints the disturbed regions from the intrinsic image and excludes them from training process, there is room for high-frequency detail to be lost depending on the quality of the intrinsic image output.
Therefore, to evaluate the impact of these potential sources of instability on the performance, we measure the generalization performance of the proposed method through evaluation on the ScanNet-NoReflection set, where there are few reflective surfaces in the scene.
As shown in~\tabref{tab:noreflection}, on average, across all metrics, Monodepth2 has a 1.47\% performance drop, HRDepth shows a 0.42\% performance improvement, and MonoViT records a 1.10\% performance drop. However, compared to the performance gains achieved by the proposed method on the reflection set, the performance drops for Monodepth2 and MonoViT are negligible, while HRDepth even shows performance gains.
This suggests that the proposed method accurately localizes reflective surfaces and achieves strong generalization performance by preserving features that are conducive to training.

\input{tables/table3}
\input{tables/table4}

\subsubsection{7-Scenes \& Booster dataset}

To further validate the effectiveness of the proposed method across various environments, we evaluate its cross-dataset generalization performance on 7-Scenes and Booster datasets. These datasets are characterized by scenes dominated by non-reflective and reflective objects (\textit{i.e.}, specular, transparent, and, mirrors), respectively.
As shown in~\tabref{tab:7scene_booster}, models trained with the proposed method demonstrate consistent performance gains on the Booster dataset, which includes a significant presence of non-Lambertian surfaces. Specifically, Monodepth2 shows an improvement of 9.96\%, HRDepth improved by 8.09\%, and MonoViT by 4.39\%, averaged across all metrics. These results indicate that the models trained with the proposed method are robust in estimating various non-Lambertian surfaces effectively.

On the other hand, on the 7-Scenes dataset, predominantly featuring opaque objects, our method achieves average performance improvements of 1.38\% on Monodepth2 and 1.80\% on MonoViT across all evaluation metrics. 
HRDepth shows an average performance degradation of 2.15\% across all metrics. However, the decrease is considered negligible compared to the significant 8.09\% improvement observed on non-Lambertian surfaces of the Booster dataset.

\subsubsection{Comparison with multi-stage training methods}

To more fairly evaluate the scalability of our proposed end-to-end training method, we propose a simple yet efficient knowledge distillation technique.
We designate the model trained with the proposed method (designated as [Baseline + Ours]) and those trained using the self-supervised method with conventional photometric error (designated as [Baseline]) as the teacher models.
Distillation is performed using a mask generated from the photometric error between each RGB and diffuse image, as defined in ~\equref{equ:reflection_mask}.
Specifically, if the photometric error for diffuse images exceeds that for RGB images by a certain margin, the depth learned by the [Baseline] method under conventional photometric error conditions is treated as pseudo-ground truth.
Conversely, if the photometric error for RGB images is significantly higher, the depth from [Baseline + Ours] is used as pseudo ground truth. (Detailed distillation process is provided in the appendix).
As shown in~\tabref{tab:multi-stage}, our proposed distillation model outperforms the self-teaching approach~\citep{poggi2020uncertainty}, which uses uncertainty-based fusion for generating pseudo ground truth, with a significant enhancement across all metrics. 
Compared to 3D Distillation~\citep{shi20233d}, our method achieves an average performance improvement of 2.57\% for Monodepth2 across all metrics, with HRDepth showing an improvement of 0.52\%, while MonoViT exhibits a performance decrease of 5.66\%.
Notably, the proposed method achieves comparable performance to 3D Distillation but achieves this without the extensive computational demands associated with 3D Distillation's approach of using uncertainty-based multi-model ensembles and 3D mesh reconstruction for pseudo ground truth generation.

\begin{table}[h!]
  \vspace{-5pt}
  \begin{center}
  \resizebox{\columnwidth}{!}{
  \begin{tabular}{@{}lcc@{}}
  \toprule
    \multirow{2}{*}{Method} & \multicolumn{2}{c}{ScanNet-Reflection Test Set} \\
    & Abs Rel $\downarrow$ & RMSE $\downarrow$ \\
    \midrule
    Monodepth2 & 0.181 & 0.521 \\
    Monodepth2 + \textit{Ours} (w/o. $\mathcal{L}_{\text{cts}}$) & 0.176 & 0.510 \\
    Monodepth2 + \textit{Ours} & \textbf{0.163} & \textbf{0.472} \\
  \bottomrule
  \end{tabular}}
  \end{center}
  \caption{Ablation study on the ScanNet-Reflection Test set.}
  \label{tab:ablation}
  \vspace{-10pt}
\end{table}

\subsubsection{Ablation study of intrinsic loss term}
In the loss terms of the intrinsic network, $\mathcal{L}_\text{recon}$ and $\mathcal{L}_\text{cross}$ serve as fundamental conditions necessary to derive meaningful intrinsic images.
To investigate the impact of the proposed contrastive loss, $\mathcal{L}_\text{cts}$, we integrate it into Monodepth2 and compare the performance changes with and without this loss.
As shown in \tabref{tab:ablation}, the experimental results show that the omitting $\mathcal{L}_\text{cts}$ causes the network to learn the diffuse image that approaches to a non-valid image, adversely affecting the accuracy of reflective region localization. This misalignment results in marginal performance improvements.


%% file: tables/table2.tex
\begin{table*}[t!]
  \begin{center}
  \resizebox{0.8\textwidth}{!}{
  \begin{tabular}{@{}lcccccccc@{}}
  \toprule
    \multirow{2}{*}{Method} & \multicolumn{7}{c}{ScanNet-NoReflection Validation Set} \\
    & Abs Rel $\downarrow$ & Sq Rel $\downarrow$ & RMSE $\downarrow$ & RMSE log $\downarrow$ & $\delta < 1.25 \uparrow$ & $\delta < 1.25^2 \uparrow$ & $\delta < 1.25^3 \uparrow$ \\
    \midrule
    Monodepth2 & \textbf{0.169} & \textbf{0.100} & \textbf{0.395} & \textbf{0.206} & \textbf{0.759} & \textbf{0.932} & \textbf{0.979} \\
    Monodepth2 + \textit{Ours} & 0.174 & 0.103 & 0.398 & 0.210 & 0.752 & 0.927 & 0.977 \\
    \midrule
    HRDepth & \textbf{0.169} & 0.102 & \textbf{0.388} & \textbf{0.202} & \textbf{0.766} & \textbf{0.933} & \textbf{0.980} \\
    HRDepth + \textit{Ours} & \textbf{0.169} & \textbf{0.097} & \textbf{0.388} & 0.205 & 0.763 & 0.932 & \textbf{0.980} \\
    \midrule
    MonoViT & \textbf{0.140} & 0.074 & \textbf{0.333} & \textbf{0.171} & \textbf{0.829} & \textbf{0.952} & 0.984 \\
    MonoViT + \textit{Ours} & 0.144 & \textbf{0.073} & 0.339 & 0.176 & 0.816 & 0.951 & \textbf{0.986} \\
  \bottomrule
  \end{tabular}}
  \end{center}
  \caption{Main results on the ScanNet-NoReflection Validation set.}
  \label{tab:noreflection}
  \vspace{-3pt}
\end{table*}

%% file: tables/table3.tex
\begin{table*}[t!]

  \begin{center}
  \resizebox{0.9\textwidth}{!}{
  \begin{tabular}{@{}c|lccccccc@{}}
  \toprule
    Dataset & Method & Abs Rel $\downarrow$ & Sq Rel $\downarrow$ & RMSE $\downarrow$ & RMSE log $\downarrow$ & $\delta < 1.25 \uparrow$ & $\delta < 1.25^2 \uparrow$ & $\delta < 1.25^3 \uparrow$ \\
    \midrule
    \multirow{6}{*}{\rotatebox[origin=c]{90}{\makecell{7-Scenes}}} & Monodepth2 & 0.210 & 0.130 & 0.445 & 0.248 & 0.656 & \textbf{0.906} & \textbf{0.974} \\
    & Monodepth2 + \textit{Ours} & \textbf{0.209} & \textbf{0.126} & \textbf{0.431} & \textbf{0.244} & \textbf{0.666} & \textbf{0.906} & 0.972 \\
    \cmidrule{2-9}
    & HRDepth & \textbf{0.193} & \textbf{0.115} & \textbf{0.421} & \textbf{0.231} & \textbf{0.682} & \textbf{0.921} & \textbf{0.982} \\
    & HRDepth + \textit{Ours} & 0.204 & 0.116 & \textbf{0.421} & 0.239 & 0.661 & 0.907 & 0.978 \\
    \cmidrule{2-9}
    & MonoViT & 0.173 & 0.093 & 0.365 & 0.201 & 0.752 & 0.945 & 0.988 \\
    & MonoViT + \textit{Ours} & \textbf{0.171} & \textbf{0.086} & \textbf{0.355} & \textbf{0.200} & \textbf{0.754} & \textbf{0.949} & \textbf{0.988} \\
  \bottomrule
  \toprule
    \multirow{6}{*}{\rotatebox[origin=c]{90}{\makecell{Booster}}} & Monodepth2 & 0.520 & 0.429 & 0.601 & 0.444 & 0.305 & 0.591 & 0.827 \\
    & Monodepth2 + \textit{Ours} & \textbf{0.471} & \textbf{0.341} & \textbf{0.536} & \textbf{0.412} & \textbf{0.334} & \textbf{0.639} & \textbf{0.861} \\
    \cmidrule{2-9}
    & HRDepth & 0.495 & 0.391 & 0.559 & 0.426 & 0.307 & 0.611 & 0.852 \\
    & HRDepth + \textit{Ours} & \textbf{0.462} & \textbf{0.315} & \textbf{0.506} & \textbf{0.404} & \textbf{0.325} & \textbf{0.645} & \textbf{0.890} \\
    \cmidrule{2-9}
    & MonoViT & 0.418 & 0.327 & 0.504 & 0.374 & 0.425 & 0.679 & 0.888 \\
    & MonoViT + \textit{Ours} & \textbf{0.396} & \textbf{0.293} & \textbf{0.482} & \textbf{0.364} & \textbf{0.426} & \textbf{0.716} & \textbf{0.909} \\
  \bottomrule
  \end{tabular}}
  \end{center}
  \caption{Main results on the 7-Scenes and Booster dataset.}
  \label{tab:7scene_booster}
\end{table*}

%% file: tables/table4.tex
\begin{table*}[t!]
  \begin{center}
  \resizebox{0.9\textwidth}{!}{
  \begin{tabular}{@{}lcccccccc@{}}
  \toprule
    \multirow{2}{*}{Backbone} & \multirow{2}{*}{Method} & \multicolumn{7}{c}{ScanNet-Reflection Validation Set} \\
    & & Abs Rel $\downarrow$ & Sq Rel $\downarrow$ & RMSE $\downarrow$ & RMSE log $\downarrow$ & $\delta < 1.25 \uparrow$ & $\delta < 1.25^2 \uparrow$ & $\delta < 1.25^3 \uparrow$ \\
    \midrule
    \multirow{3}{*}{Monodepth2} & Self-teaching & 0.192 & 0.188 & 0.548 & 0.233 & 0.764 & 0.920 & 0.967 \\
    & 3D Distillation & 0.156 & 0.093 & 0.442 & 0.191 & 0.786 & 0.943 & 0.987 \\
    \cmidrule{2-9}
    & \textit{Ours (distillation)} & \textbf{0.148} & \textbf{0.089} & \textbf{0.429} & \textbf{0.185} & \textbf{0.798} & \textbf{0.950} & \textbf{0.989} \\
    \midrule
    \multirow{3}{*}{HRDepth} & Self-teaching & 0.202 & 0.208 & 0.565 & 0.243 & 0.756 & 0.914 & 0.964 \\
    & 3D Distillation & 0.153 & \textbf{0.090} & \textbf{0.430} & 0.188 & 0.789 & 0.948 & \textbf{0.989} \\
    \cmidrule{2-9}
    & \textit{Ours (distillation)} & \textbf{0.150} & \textbf{0.090} & \textbf{0.430} & \textbf{0.187} & \textbf{0.798} & \textbf{0.949} & 0.988 \\
    \midrule
    \multirow{3}{*}{MonoViT} & Self-teaching & 0.176 & 0.195 & 0.537 & 0.224 & 0.823 & 0.930 & 0.963 \\
    & 3D Distillation & \textbf{0.126} & \textbf{0.068} & \textbf{0.367} & \textbf{0.159} & 0.851 & \textbf{0.965} & \textbf{0.991} \\
    \cmidrule{2-9}
    & \textit{Ours (distillation)} & 0.129 & 0.083 & 0.406 & 0.166 & \textbf{0.856} & 0.960 & 0.989 \\
  \bottomrule
  \end{tabular}}
  \end{center}
  \caption{Comparison between the proposed methods and Multi-stage methods on ScanNet-Reflection Validation set.}
  \label{tab:multi-stage}
  \vspace{-5pt}
\end{table*}

%% file: 5_appendix.tex
\twocolumn[
\section*{{\large Supplementary Material for}\\\vspace{5pt}{\LARGE Intrinsic Image Decomposition for Robust Self-supervised Monocular Depth\\\vspace{3pt}Estimation on Reflective Surfaces}}
\vspace{9pt}
]

\section{Detailed Distillation Process}
In this section, we provide a detailed description of our end-to-end training method as outlined in the main manuscripts.
Our strategy involves a novel adaption of the fusion-distillation scheme proposed in~\citet{shi20233d}, which we have modified to accommodate a multi-stage training process.
This revised method leverages two pre-trained networks as teacher models; one is trained using the traditional self-supervised method, while the other is trained by the proposed reflection-aware methods.
We obtain the depth map $D_\text{org}$ from the conventional self-supervised training method~\cite{godard2019digging}, and the depth map $D_\text{refl.}$ from the proposed training method, respectively.
$D_\text{org}$ excels in capturing high-frequency details, such as small objects or edges in non-reflective regions, due to consistent photometric error minimization.
On the other hand, $D_\text{refl.}$ achieves robust depth accuracy in reflective regions, where $D_\text{org}$ typically suffers from significant depth quality degradation.
To leverage the strengths of both maps, we fuse these depth maps into a single comprehensive depth map $D_\text{pseudo}$, utilizing a pixel-level reflective mask $M_R$ as follows:
\begin{equation}
\begin{split}
    \label{equ:pseudo_depth}
    D_\text{pseudo} = M_R \odot D_\text{org} + (1 - M_R) \odot D_\text{refl.},
\end{split}
\end{equation}
where $M_R$ is obtained from the depth map and intrinsic images produced by our method, as detailed in Equation 8-10 in our main manuscript.

To address the issue of infinite holes exceeding the actual residual component size, as depicted in Figure 3 of the main manuscript, we have modified Equation 10. A margin term, $\delta = 0.1$, has been introduced to tighten the criteria of selecting non-reflective regions as follows:
\begin{align}
\begin{split}
\label{equ:revised_reflection_mask}
    &M_{R}[u, v] = 
    \begin{cases}
    0, \text{ if }z_{L}[u,v] < z_{I}[u,v] ~+~\delta\\
    1, \text{ else }
    \end{cases},~\text{where}\\
    z_{I} &= \sum_{u,v}\sqrt{(E_{I}[u,v]-\mu_{I})\mathbf{\sigma}_{I}^{-1}(E_{I}[u,v]-\mu_{I})^T},\\
    z_{L} &= \sum_{u,v}\sqrt{(E_{L}[u,v]-\mu_{L})\sigma_{L}^{-1}(E_{L}[u,v]-\mu_{L})^T}.
\end{split}
\end{align}
Given that $M_R$ assigns a value of 0 to reflective regions and 1 to non-reflective regions, this mask effectively mitigates issues associated with reflective surfaces while preserving high-frequency details.
To distillate the pseudo-depth label to the student network, we use the log-scale L1 loss as follows:
\begin{equation}
\begin{split}
    \mathcal{L}_\text{distill} = |\log \hat{D} - \log D_\text{pseudo}|,
\end{split}
\end{equation}
where $\hat{D}$ represents the estimated depth from the student network.
This distillation approach achieves a balanced and accurate depth prediction across various surface types.

\input{tables/apx_table1}
\input{tables/apx_table2}


\section{Comprehensive Comparison of Our End-to-End and Distillation Methods}
To demonstrate the effectiveness of our distillation method, we evaluate the performance of network trained using our distillation method, \textit{Ours (distillation)}, against the proposed end-to-end method, \textit{Ours},  across all datasets and baselines.
As shown in \tabref{tab:apx_table1}, our simple yet distillation method consistently boosts the performance with significant margins compared to \textit{Ours}.
Notably, our distillation method shows considerable improvements in depth performance on non-reflective datasets (\textit{i.e.}, ScanNet-NoReflection Validation, 7-Scenes).
The consistent enhancement of our distillation method on both non-reflective and reflective surfaces underscores the efficacy of the distillation mask $M_R$ in accurately identifying per-pixel reflective regions.

\section{Qualitative Comparison: \\``Ours" v.s. ``Ours (distillation)"}
\begin{figure*}[t!]
\centering
\includegraphics[width=0.90\textwidth]{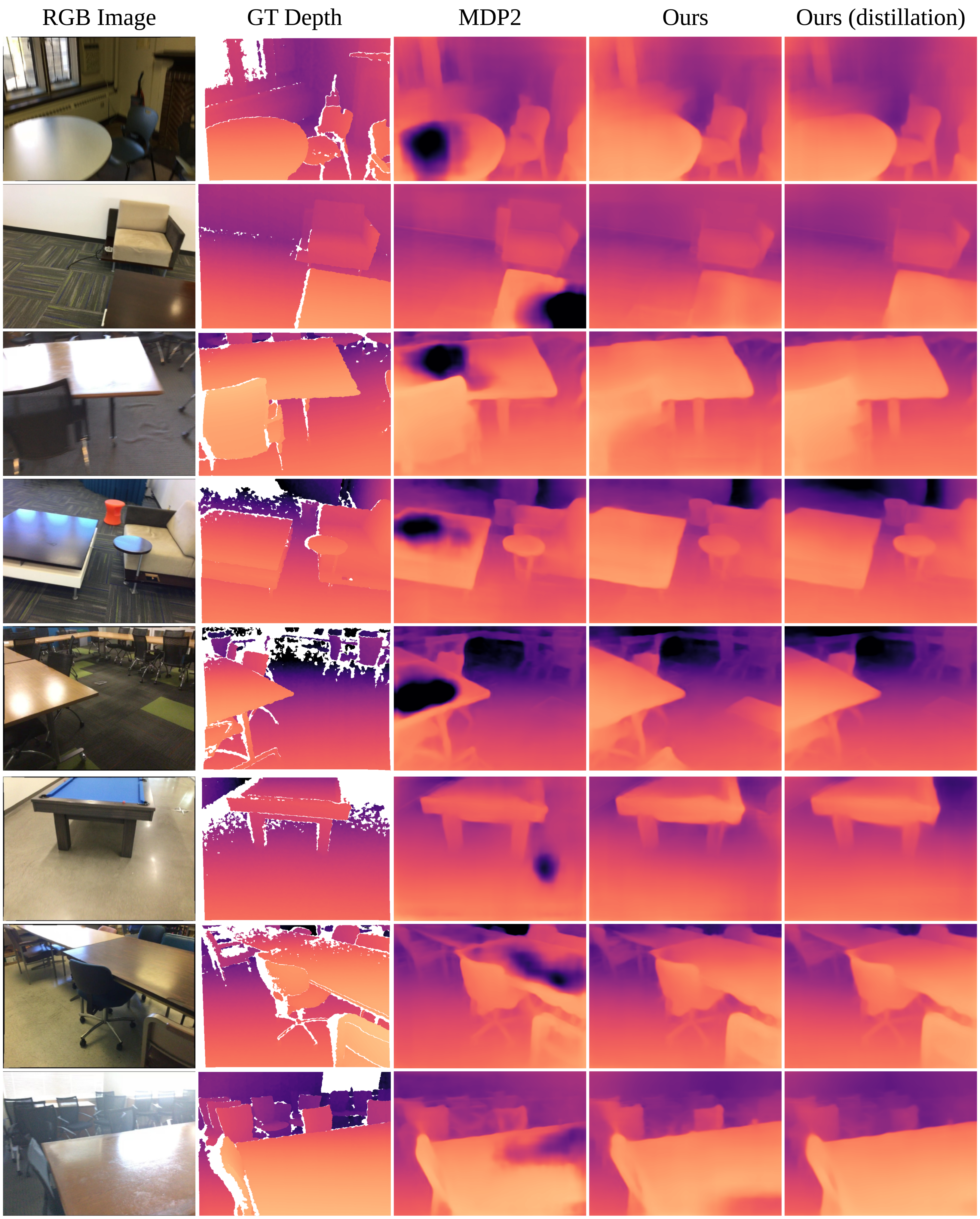} 
\caption{Qualitative comparison of ``Ours", ``Ours (distillation)", and Monodepth2 (denoted as MDP2).}
\label{fig:supp_qualitative}
\end{figure*}

We provide the additional qualitative results of our method with and without the distillation.
As shown in \figref{fig:supp_qualitative}, ``Ours (distillation)" effectively enhances the high-frequency details on non-reflective surfaces while preserving the depth quality on reflective surfaces compared to ``Ours".

\section{Quantitative Comparison: \\ Our Distillation Method v.s. 3D distillation.}
Lastly, we summarize the additional quantitative comparison between our distillation method and 3D distillation~\cite{shi20233d} on the ScanNet triplet splits in~\tabref{tab:apx_table2}.

\section{Limitation and Future Work}
Our method can effectively improve depth performance in reflective regions, such as specular highlights from strong light sources or in areas with low residual intensity.
However, the challenge of accurately estimating depth for hard-case non-Lambertian objects, such as transparent objects or fully reflective objects (i.e., mirrors), remains.
Additionally, this paper does not address occlusion-related issues like edge fattening, as they fall outside our primary focus.
In our intrinsic residual formulation, we empirically utilize a 3-channel diffuse and grayscale residual to maintain training stability and simplicity.
However, based on dichromatic modeling~\citep{tominaga1994dichromatic}, a three-channel configuration for the residual could potentially be more effective, considering the chromatic differences between the specular and diffuse components.
Furthermore, most prior intrinsic image decomposition methods adopt loss functions enriched with prior knowledge to enhance the accuracy of the intrinsic images.
We expect that these advancements in intrinsic residual modeling will further enhance the accuracy of depth estimation in our method.

%% file: tables/apx_table1.tex
\begin{table*}[!ht]
  \caption{Quantitative comparison (full version) of the ``\textit{Ours}" and ``\textit{Ours (distillation)}"}
  \label{tab:apx_table1}
  \begin{center}
  \resizebox{0.75\textwidth}{!}{
  \begin{tabular}{@{}cllccccccc@{}}
  \toprule
    Dataset & Backbone & Method & Abs Rel $\downarrow$ & Sq Rel $\downarrow$ & RMSE $\downarrow$ & RMSE log $\downarrow$ & $\delta < 1.25 \uparrow$ & $\delta < 1.25^2 \uparrow$ & $\delta < 1.25^3 \uparrow$ \\
    \midrule
    \multirow{6}{*}{\rotatebox[origin=c]{90}{\makecell{ScanNet-Refl.\\Test set}}} & \multirow{2}{*}{Monodepth2} & \textit{Ours} & 0.163 & 0.103 & 0.472 & 0.205 & 0.757 & 0.940 & 0.986 \\
    & & \textit{Ours (distillation)} & \textbf{0.153} & \textbf{0.092} & \textbf{0.444} & \textbf{0.192} & \textbf{0.785} & \textbf{0.949} & \textbf{0.989} \\
    \cmidrule{2-10}
    & \multirow{2}{*}{HRDepth} & \textit{Ours} & 0.165 & 0.104 & 0.471 & 0.206 & 0.752 & 0.934 & 0.984 \\
    & & \textit{Ours (distillation)} & \textbf{0.151} & \textbf{0.090} & \textbf{0.439} & \textbf{0.190} & \textbf{0.778} & \textbf{0.946} & \textbf{0.989} \\
    \cmidrule{2-10}
    & \multirow{2}{*}{MonoViT} & \textit{Ours} & 0.142 & 0.093 & 0.421 & 0.180 & 0.815 & 0.957 & \textbf{0.985} \\
    & & \textit{Ours (distillation)} & \textbf{0.141} & \textbf{0.092} & \textbf{0.415} & \textbf{0.175} & \textbf{0.830} & \textbf{0.952} & \textbf{0.985} \\
  \toprule
    \multirow{6}{*}{\rotatebox[origin=c]{90}{\makecell{ScanNet-Refl.\\Val set}}} & \multirow{2}{*}{Monodepth2} & \textit{Ours} & 0.158 & 0.100 & 0.462 & 0.200 & 0.769 & 0.939 & 0.986 \\
    & & \textit{Ours (distillation)} & \textbf{0.148} & \textbf{0.089} & \textbf{0.429} & \textbf{0.185} & \textbf{0.798} & \textbf{0.950} & \textbf{0.989} \\
    \cmidrule{2-10}
    & \multirow{2}{*}{HRDepth} & \textit{Ours} & 0.160 & 0.102 & 0.463 & 0.201 & 0.773 & 0.939 & 0.986 \\
    & & \textit{Ours (distillation)} & \textbf{0.150} & \textbf{0.090} & \textbf{0.430} & \textbf{0.187} & \textbf{0.798} & \textbf{0.949} & \textbf{0.988} \\
    \cmidrule{2-10}
    & \multirow{2}{*}{MonoViT} & \textit{Ours} & 0.139 & 0.106 & 0.446 & 0.179 & 0.840 & 0.953 & 0.984 \\
    & & \textit{Ours (distillation)} & \textbf{0.129} & \textbf{0.083} & \textbf{0.406} & \textbf{0.166} & \textbf{0.856} & \textbf{0.960} & \textbf{0.989} \\
  \toprule
    \multirow{6}{*}{\rotatebox[origin=c]{90}{\makecell{ScanNet-NoRefl.\\Val set}}} & \multirow{2}{*}{Monodepth2} & \textit{Ours} & 0.174 & 0.103 & 0.398 & 0.210 & 0.752 & 0.927 & 0.977 \\
    & & \textit{Ours (distillation)} & \textbf{0.164} & \textbf{0.090} & \textbf{0.370} & \textbf{0.196} & \textbf{0.775} & \textbf{0.937} & \textbf{0.983} \\
    \cmidrule{2-10}
    & \multirow{2}{*}{HRDepth} & \textit{Ours} & 0.169 & 0.097 & 0.388 & 0.205 & 0.763 & 0.932 & 0.980 \\
    & & \textit{Ours (distillation)} & \textbf{0.160} & \textbf{0.088} & \textbf{0.365} & \textbf{0.194} & \textbf{0.783} & \textbf{0.938} & \textbf{0.983} \\
    \cmidrule{2-10}
    & \multirow{2}{*}{MonoViT} & \textit{Ours} & 0.144 & 0.073 & 0.339 & 0.176 & 0.816 & 0.951 & \textbf{0.986} \\
    & & \textit{Ours (distillation)} & \textbf{0.138} & \textbf{0.069} & \textbf{0.321} & \textbf{0.167} & \textbf{0.833} & \textbf{0.955} & \textbf{0.986} \\
  \toprule
    \multirow{6}{*}{\rotatebox[origin=c]{90}{\makecell{7-Scenes}}} & \multirow{2}{*}{Monodepth2} & \textit{Ours} & 0.209 & 0.126 & 0.431 & 0.244 & 0.666 & 0.906 & 0.972 \\
    & & \textit{Ours (distillation)} & \textbf{0.195} & \textbf{0.105} & \textbf{0.397} & \textbf{0.228} & \textbf{0.689} & \textbf{0.922} & \textbf{0.981} \\
    \cmidrule{2-10}
    & \multirow{2}{*}{HRDepth} & \textit{Ours} & 0.204 & 0.116 & 0.421 & 0.239 & 0.661 & 0.907 & 0.978 \\
    & & \textit{Ours (distillation)} & \textbf{0.187} & \textbf{0.097} & \textbf{0.388} & \textbf{0.222} & \textbf{0.696} & \textbf{0.924} & \textbf{0.985} \\
    \cmidrule{2-10}
    & \multirow{2}{*}{MonoViT} & \textit{Ours} & 0.171 & 0.086 & 0.355 & 0.200 & 0.754 & 0.949 & 0.988 \\
    & & \textit{Ours (distillation)} & \textbf{0.161} & \textbf{0.075} & \textbf{0.328} & \textbf{0.187} & \textbf{0.782} & \textbf{0.955} & \textbf{0.990} \\
  \toprule
    \multirow{6}{*}{\rotatebox[origin=c]{90}{\makecell{Booster}}} & \multirow{2}{*}{Monodepth2} & \textit{Ours} & 0.471 & 0.341 & 0.536 & 0.412 & 0.334 & 0.639 & 0.861 \\
    & & \textit{Ours (distillation)} & \textbf{0.441} & \textbf{0.291} & \textbf{0.487} & \textbf{0.386} & \textbf{0.355} & \textbf{0.666} & \textbf{0.894} \\
    \cmidrule{2-10}
    & \multirow{2}{*}{HRDepth} & \textit{Ours} & 0.462 & 0.315 & 0.506 & 0.404 & 0.325 & 0.645 & 0.890 \\
    & & \textit{Ours (distillation)} & \textbf{0.432} & \textbf{0.285} & \textbf{0.483} & \textbf{0.389} & \textbf{0.346} & \textbf{0.673} & \textbf{0.891} \\
    \cmidrule{2-10}
    & \multirow{2}{*}{MonoViT} & \textit{Ours} & \textbf{0.396} & 0.293 & 0.482 & 0.364 & \textbf{0.426} & \textbf{0.716} & 0.909 \\
    & & \textit{Ours (distillation)} & 0.404 & \textbf{0.282} & \textbf{0.473} & \textbf{0.363} & 0.414 & 0.692 & \textbf{0.923} \\
  \bottomrule
  \end{tabular}}
  \end{center}
\end{table*}

%% file: tables/apx_table2.tex
\begin{table*}[!ht]
  \caption{Quantitative comparison of the proposed methods and Multi-stage methods on ScanNet dataset.}
  \label{tab:apx_table2}
  \begin{center}
  \resizebox{0.75\textwidth}{!}{
  \begin{tabular}{@{}cllccccccc@{}}
  \toprule
    Split & Backbone & Method & Abs Rel $\downarrow$ & Sq Rel $\downarrow$ & RMSE $\downarrow$ & RMSE log $\downarrow$ & $\delta < 1.25 \uparrow$ & $\delta < 1.25^2 \uparrow$ & $\delta < 1.25^3 \uparrow$ \\
    \midrule
    \multirow{9}{*}{\rotatebox[origin=c]{90}{\makecell{ScanNet-Refl.\\Test set}}} & \multirow{3}{*}{Monodepth2} & Self-teaching & 0.179 & 0.146 & 0.502 & 0.218 & 0.750 & 0.938 & 0.980 \\
    & & 3D Distillation & 0.156 & 0.096 & 0.459 & 0.195 & 0.766 & 0.945 & 0.988 \\
    \cmidrule{3-10}
    & & \textit{Ours (distillation)} & \textbf{0.153} & \textbf{0.092} & \textbf{0.444} & \textbf{0.192} & \textbf{0.785} & \textbf{0.949} & \textbf{0.989} \\
    \cmidrule{2-10}
    & \multirow{3}{*}{HRDepth} & Self-teaching & 0.175 & 0.145 & 0.492 & 0.215 & 0.757 & 0.936 & 0.982 \\
    & & 3D Distillation & 0.152 & \textbf{0.089} & 0.451 & \textbf{0.190} & 0.771 & \textbf{0.956} & \textbf{0.990} \\
    \cmidrule{3-10}
    & & \textit{Ours (distillation)} & \textbf{0.151} & 0.090 & \textbf{0.439} & \textbf{0.190} & \textbf{0.778} & 0.946 & 0.989 \\
    \cmidrule{2-10}
    & \multirow{3}{*}{MonoViT} & Self-teaching & 0.151 & 0.130 & 0.439 & 0.191 & 0.837 & 0.950 & 0.978 \\
    & & 3D Distillation & \textbf{0.127} & \textbf{0.069} & \textbf{0.379} & \textbf{0.162} & \textbf{0.846} & \textbf{0.961} & \textbf{0.992} \\
    \cmidrule{3-10}
    & & \textit{Ours (distillation)} & 0.141 & 0.092 & 0.415 & 0.175 & 0.830 & 0.952 & 0.985 \\
  \midrule    
    \multirow{9}{*}{\rotatebox[origin=c]{90}{\makecell{ScanNet-Refl.\\Val set}}} & \multirow{3}{*}{Monodepth2} & Self-teaching & 0.192 & 0.188 & 0.548 & 0.233 & 0.764 & 0.920 & 0.967 \\
    & & 3D Distillation & 0.156 & 0.093 & 0.442 & 0.191 & 0.786 & 0.943 & 0.987 \\
    \cmidrule{3-10}
    & & \textit{Ours (distillation)} & \textbf{0.148} & \textbf{0.089} & \textbf{0.429} & \textbf{0.185} & \textbf{0.798} & \textbf{0.950} & \textbf{0.989} \\
    \cmidrule{2-10}
    & \multirow{3}{*}{HRDepth} & Self-teaching & 0.202 & 0.208 & 0.565 & 0.243 & 0.756 & 0.914 & 0.964 \\
    & & 3D Distillation & 0.153 & \textbf{0.090} & \textbf{0.430} & 0.188 & 0.789 & 0.948 & \textbf{0.989} \\
    \cmidrule{3-10}
    & & \textit{Ours (distillation)} & \textbf{0.150} & \textbf{0.090} & \textbf{0.430} & \textbf{0.187} & \textbf{0.798} & \textbf{0.949} & 0.988 \\
    \cmidrule{2-10}
    & \multirow{3}{*}{MonoViT} & Self-teaching & 0.176 & 0.195 & 0.537 & 0.224 & 0.823 & 0.930 & 0.963 \\
    & & 3D Distillation & \textbf{0.126} & \textbf{0.068} & \textbf{0.367} & \textbf{0.159} & 0.851 & \textbf{0.965} & \textbf{0.991} \\
    \cmidrule{3-10}
    & & \textit{Ours (distillation)} & 0.129 & 0.083 & 0.406 & 0.166 & \textbf{0.856} & 0.960 & 0.989 \\
  \midrule
    \multirow{9}{*}{\rotatebox[origin=c]{90}{\makecell{ScanNet-NoRefl.\\Val set}}} & \multirow{3}{*}{Monodepth2} & Self-teaching & 0.161 & 0.090 & 0.375 & 0.196 & 0.777 & 0.939 & 0.981 \\
    & & 3D Distillation & \textbf{0.159} & \textbf{0.087} & 0.373 & \textbf{0.195} & \textbf{0.779} & \textbf{0.941} & \textbf{0.983} \\
    \cmidrule{3-10}
    & & \textit{Ours (distillation)} & 0.164 & 0.090 & \textbf{0.370} & 0.196 & 0.775 & 0.937 & \textbf{0.983} \\
    \cmidrule{2-10}
    & \multirow{3}{*}{HRDepth} & Self-teaching & 0.160 & 0.089 & 0.367 & 0.192 & 0.784 & 0.941 & 0.982 \\
    & & 3D Distillation & \textbf{0.158} & \textbf{0.086} & \textbf{0.365} & \textbf{0.190} & \textbf{0.786} & \textbf{0.942} & \textbf{0.983} \\
    \cmidrule{3-10}
    & & \textit{Ours (distillation)} & 0.160 & 0.088 & \textbf{0.365} & 0.194 & 0.783 & 0.938 & \textbf{0.983} \\
    \cmidrule{2-10}
    & \multirow{3}{*}{MonoViT} & Self-teaching & 0.134 & 0.068 & 0.317 & 0.164 & \textbf{0.840} & \textbf{0.956} & \textbf{0.987} \\
    & & 3D Distillation & \textbf{0.133} & \textbf{0.065} & \textbf{0.311} & \textbf{0.162} & 0.838 & \textbf{0.956} & \textbf{0.987} \\
    \cmidrule{3-10}
    & & \textit{Ours (distillation)} & 0.138 & 0.069 & 0.321 & 0.167 & 0.833 & 0.955 & 0.986 \\
  \bottomrule

  \end{tabular}}
  \end{center}
\end{table*}